\documentclass[10pt,conference]{IEEEtran}

\newif\ifieee
\ieeefalse

\newif\ifarxiv
\arxivtrue

\ifieee
    \ifarxiv
        \PackageError{Configuration}
        {IEEE and arXiv cannot both be enabled}
        {Set either \string\ieeefalse\space or \string\arxivfalse.}
    \fi
\fi

\IEEEoverridecommandlockouts

\ifieee
\IEEEpubid{\makebox[\columnwidth]{979-8-3195-0255-1/26/\$31.00~\copyright2026 IEEE \hfill} \hspace{\columnsep}\makebox[\columnwidth]{ }}
\fi

\usepackage{multirow}
\usepackage{makecell}
\usepackage{arydshln}
\usepackage[normalem]{ulem}
\usepackage{balance}
\usepackage{newtxtext}
\usepackage{comment}
\usepackage{stfloats}
\usepackage{amssymb}

\usepackage{tabularx}
\usepackage{makecell}

\usepackage[sort,compress]{cite}

\ifCLASSINFOpdf
   \usepackage[pdftex]{graphicx}
   \graphicspath{{figs/}}
   \DeclareGraphicsExtensions{.pdf,.jpeg,.png}
\else
   \usepackage[dvips]{graphicx}
   \graphicspath{{../figs/}}
   \DeclareGraphicsExtensions{.eps}
\fi

\usepackage[cmex10]{amsmath}
\usepackage{textcomp}
\usepackage{gensymb}
\usepackage{amsthm}

\usepackage{algorithmic}

\usepackage{array}

\ifCLASSOPTIONcompsoc
  \usepackage[caption=false,font=normalsize,labelfont=sf,textfont=sf,farskip=0pt]{subfig}
\else
  \usepackage[caption=false,font=footnotesize,farskip=0pt]{subfig}
\fi

\usepackage[acronym]{glossaries}
\usepackage{xspace}
\usepackage[hyphens]{url}
\usepackage{booktabs}
\usepackage[dvipsnames]{xcolor}
\usepackage[utf8]{inputenc}
\usepackage[T1]{fontenc}

\ifieee
\else
\usepackage[colorlinks=true,allcolors=black]{hyperref} %
\fi

\usepackage[capitalise]{cleveref} %

\newacronymstyle{long-short-br}
{%
  \GlsUseAcrEntryDispStyle{long-short}%
}%
{%
  \GlsUseAcrStyleDefs{long-short}%
}
\setacronymstyle{long-short-br}

\usepackage{transparent}
\usepackage{tikz}
\ifarxiv
    \newcommand\copyrighttext{%
      \scriptsize Accepted for presentation at SIBGRAPI 2026. The final published version will be available on IEEE~Xplore.}
    \newcommand\copyrightnotice{%
    \begin{tikzpicture}[remember picture,overlay]
    \node[anchor=south,yshift=30pt,xshift=0pt] at (current page.south) {\fbox{\transparent{0.85}\parbox{\dimexpr0.6\textwidth-\fboxsep-\fboxrule\relax}{\copyrighttext}}};
    \end{tikzpicture}%
    }
\else
\fi

\newif\iffinal
\finaltrue

\iffinal
\else
\usepackage[switch]{lineno}
\fi

\newcommand*{\ADdel}[2][]{\textcolor{red}{\textbf{\ifthenelse{\equal{#1}{}}{}{}}}} %
\newcommand*{\AD}[2][]{{\textbf{\ifthenelse{\equal{#1}{}}{}{}}#2}} %
\newcommand*{\DM}[2][]{\textcolor{green}{[\textbf{\ifthenelse{\equal{#1}{}}{DM}{DM(#1)}}: #2]}}
\newcommand*{\GL}[2][]{\textcolor{brown}{[\textbf{\ifthenelse{\equal{#1}{}}{GL}{GL(#1)}}: #2]}}
\newcommand*{\BB}[2][]{\textcolor{orange}{[\textbf{\ifthenelse{\equal{#1}{}}{BB}{BB(#1)}}: #2]}}

\newcounter{fncounter}
\begin{document}

\iffinal
\newcommand{\dataset}{\texttt{UFPR-PEs}\xspace}
\newcommand{\urlDataset}{\url{https://github.com/UFPR-IPASP-PR/ufpr-pes}}
\else
\newcommand{\dataset}{\texttt{XXXX-XXx}\xspace}
\newcommand{\urlDataset}{\textit{[hidden for review]}}
\fi

\title{\dataset: A Brazilian Face Recognition Benchmark with Self-Declared Race/Color Labels}

\iffinal

\author{
\IEEEauthorblockN{Alexandre Diano\IEEEauthorrefmark{1}, Bernardo Biesseck\IEEEauthorrefmark{1}\IEEEauthorrefmark{2}, 
Gabriel Polo\IEEEauthorrefmark{1}, Vinicius Gregorio\IEEEauthorrefmark{1},\\Laura Lopes\IEEEauthorrefmark{1},  Diego Addan\IEEEauthorrefmark{1} and David Menotti\IEEEauthorrefmark{1}}
\IEEEauthorblockA{
\IEEEauthorrefmark{1}Department of Informatics, Federal University of Paraná, Curitiba, Brazil \\
\IEEEauthorrefmark{2}Federal Institute of Mato Grosso (IFMT), Pontes e Lacerda, Mato Grosso, Brazil \\
\{adsouza, ghp24, vgf24, lslopes, diego, menotti\}@inf.ufpr.br, bernardo.biesseck@ifmt.edu.br \\
}
}

\else
  \author{Paper ID: 239 \\[9ex]}
  \linenumbers
\fi

\maketitle

\ifarxiv
    \copyrightnotice
\else
\fi

\newacronym{arcface}{ArcFace}{Additive Angular Margin Loss for Deep Face Recognition}
\newacronym{fr}{FR}{Face Recognition}
\newacronym{ibge}{IBGE}{Brazilian Institute of Geography and Statistics}
\newacronym{ser}{SER}{Subgroup Error Rate}
\newacronym{tar}{TAR}{True Accept Rate}
\newacronym{tse}{TSE}{Superior Electoral Court}

\ifieee
\vspace{-3.575mm}
\else
\vspace{-3.575mm}
\fi

\begin{abstract}
While face recognition systems are widely deployed, ensuring their demographic reliability and robustness under uncontrolled visual conditions remains a critical challenge. To bridge this gap, we present \dataset, a benchmark for face recognition bias evaluation using public videos of elected Brazilian politicians annotated with official self-declared race/color categories. The dataset adopts the Brazilian census taxonomy, including the \textit{parda} category, which has no direct equivalent in the U.S.- or Europe-centric schemas commonly used in prior benchmarks. Our benchmark is built from compressed public video and preserves difficult samples so that performance can be analyzed under realistic conditions. We describe the construction pipeline, report dataset statistics, and evaluate face recognition performance across verification and (closed- and open-set) identification settings, including subgroup analysis by \ADdel{ethnic group}\AD{race/color} and difficulty level. The results show that recognition performance varies substantially with image quality, and that subgroup gaps must be interpreted jointly with visual difficulty rather than in isolation. Overall, \dataset provides a reproducible and demographically grounded setting for studying face recognition bias under challenging public video conditions.
\end{abstract}

\IEEEpeerreviewmaketitle

\section{Introduction}
\label{sec:introduction}

Face recognition is a central problem in computer vision, with applications ranging from identity verification to media analysis and forensic investigation. As these systems are deployed in more diverse and consequential settings, their performance must be examined not only in terms of average accuracy, but also in terms of demographic reliability and robustness under challenging visual conditions. Prior auditing studies have shown that recognition errors can vary substantially across demographic groups, especially when evaluation moves beyond curated image settings~\cite{buolamwini2018gender,raji2019actionable}. This makes benchmark design a core issue in fair face recognition research.

\begin{figure}[!thb]
    \centering
    \captionsetup[subfigure]{labelformat=empty,captionskip=2pt}
    
    \resizebox{0.90\linewidth}{!}{%
        \subfloat[]{%
            \includegraphics[height=15.8ex]{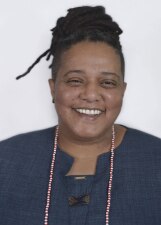}\hspace{0.5em}%
            \includegraphics[height=15.8ex]{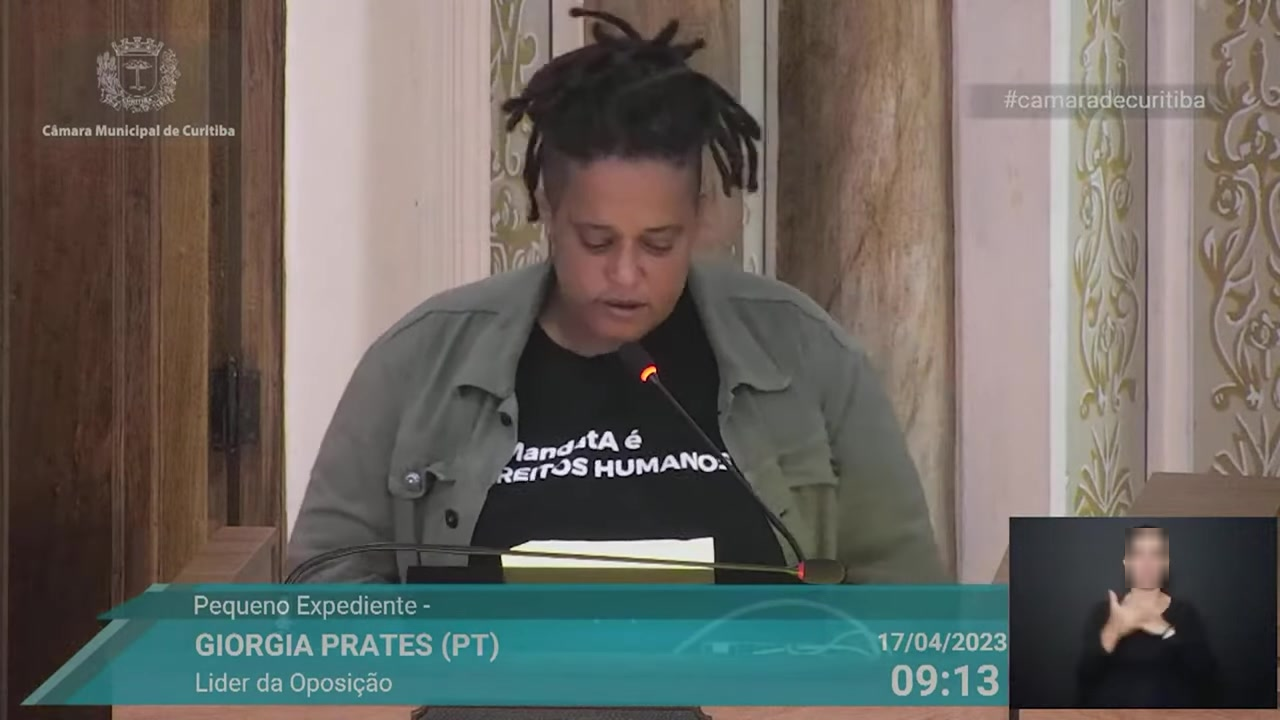}\hspace{0.5em}%
            \includegraphics[height=15.8ex]{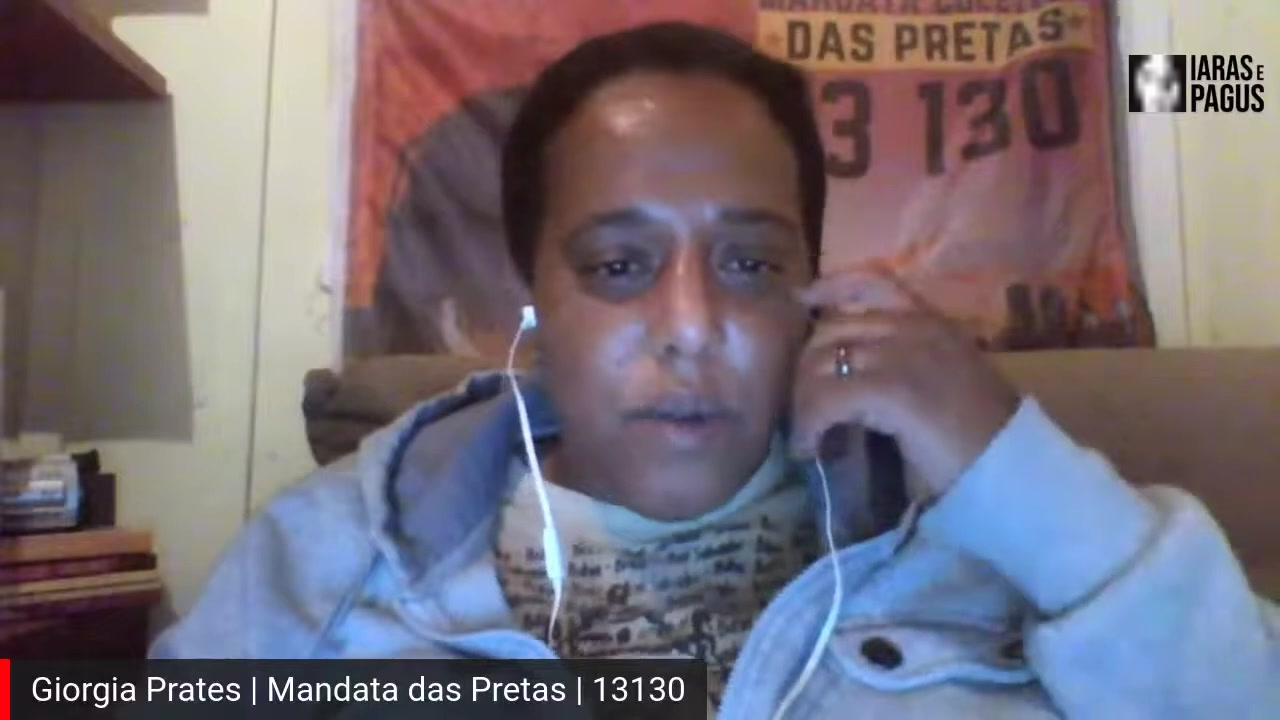}\hspace{0.5em}%
            \includegraphics[height=15.8ex]{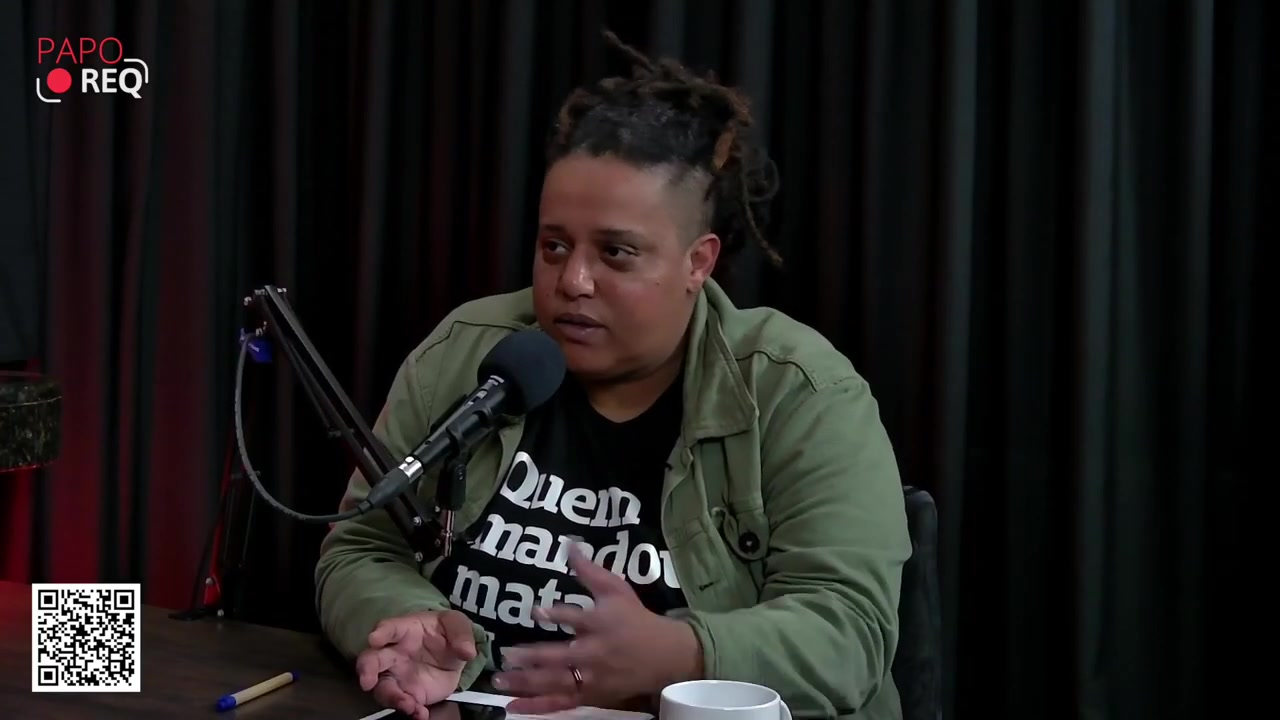}%
        }%
    }
    
    \resizebox{0.90\linewidth}{!}{%
        \subfloat[]{%
            \includegraphics[height=15.8ex]{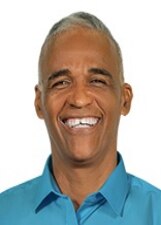}\hspace{0.5em}%
            \includegraphics[height=15.8ex]{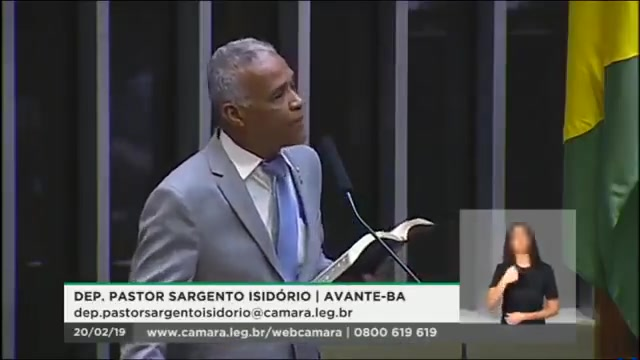}\hspace{0.5em}%
            \includegraphics[height=15.8ex]{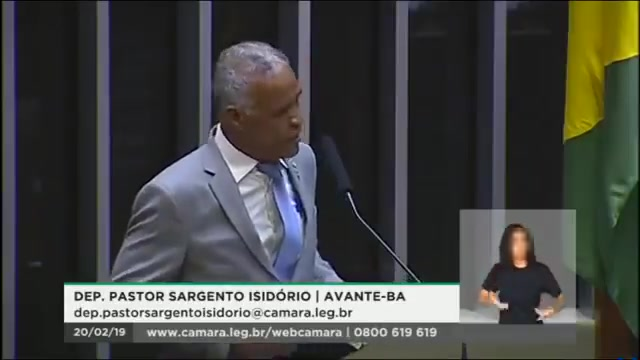}\hspace{0.5em}%
            \includegraphics[height=15.8ex]{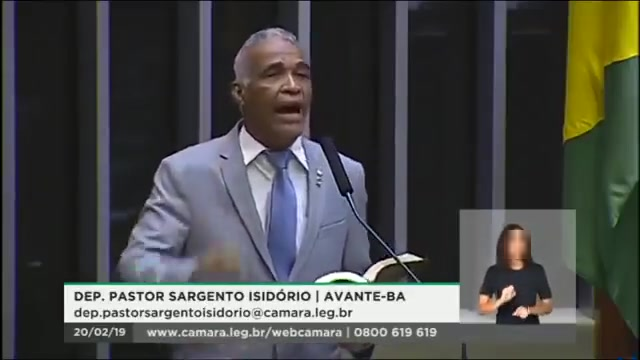}%
        }%
    }
    
    \resizebox{0.90\linewidth}{!}{%
        \subfloat[]{%
            \includegraphics[height=15.8ex]{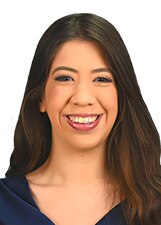}\hspace{0.5em}%
            \includegraphics[height=15.8ex]{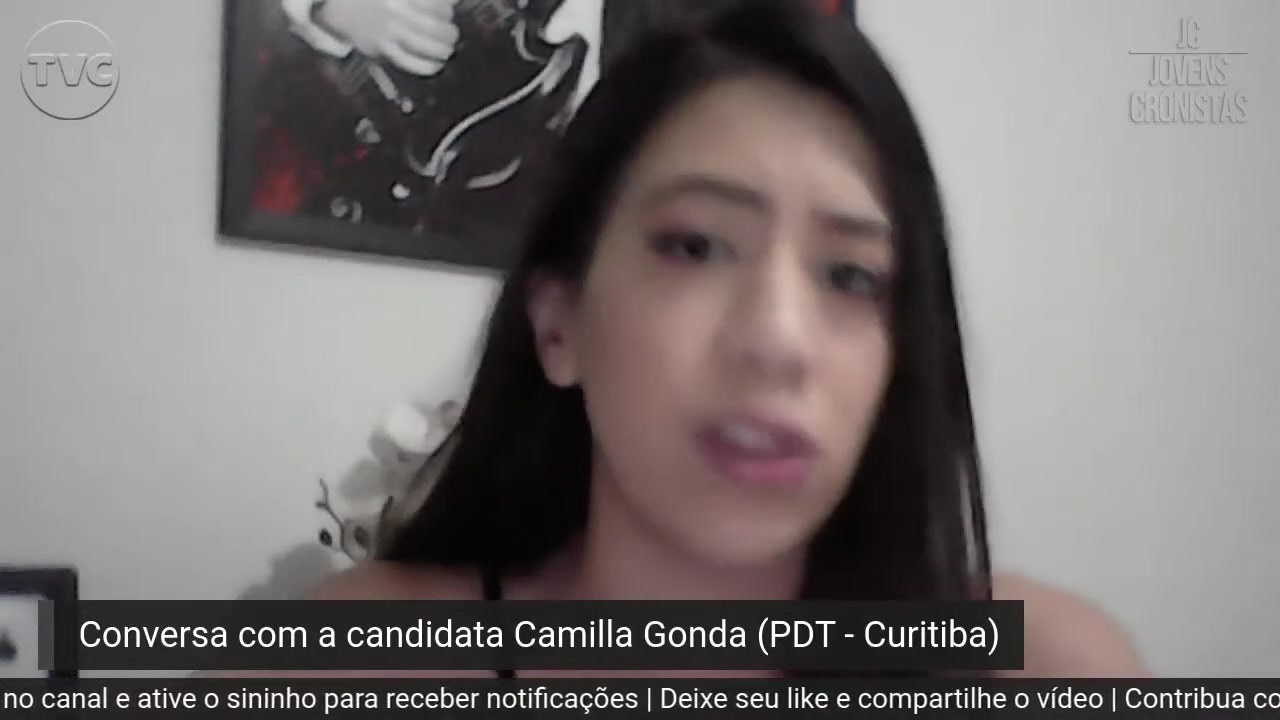}\hspace{0.5em}%
            \includegraphics[height=15.8ex]{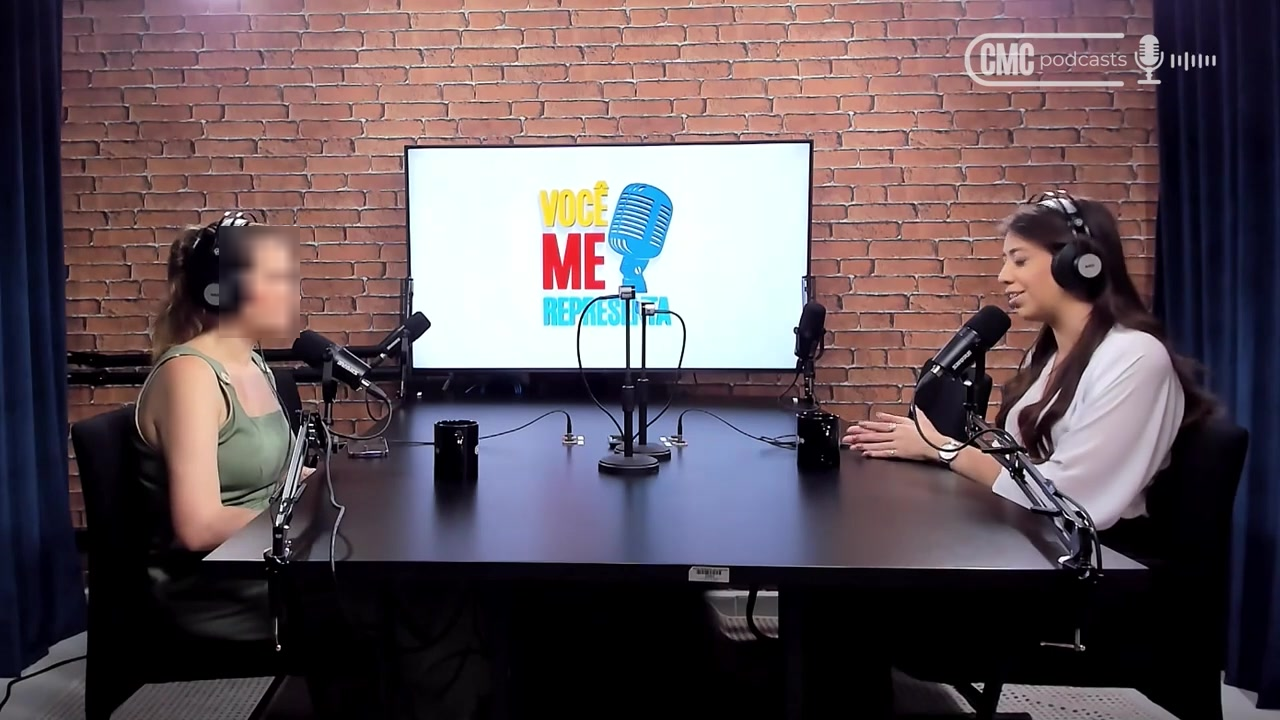}\hspace{0.5em}%
            \includegraphics[height=15.8ex]{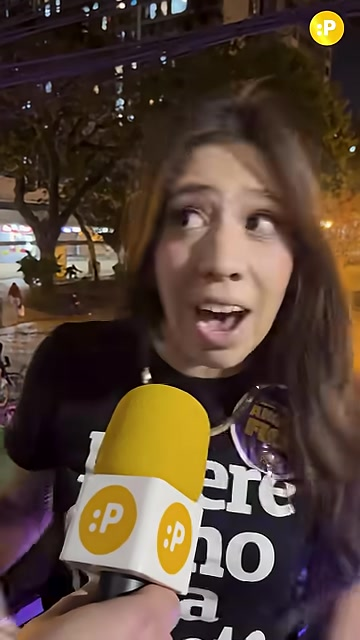}%
        }%
    }
    
    \resizebox{0.90\linewidth}{!}{%
        \subfloat[]{%
            \includegraphics[height=15.8ex]{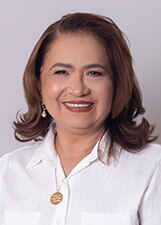}\hspace{0.5em}%
            \includegraphics[height=15.8ex]{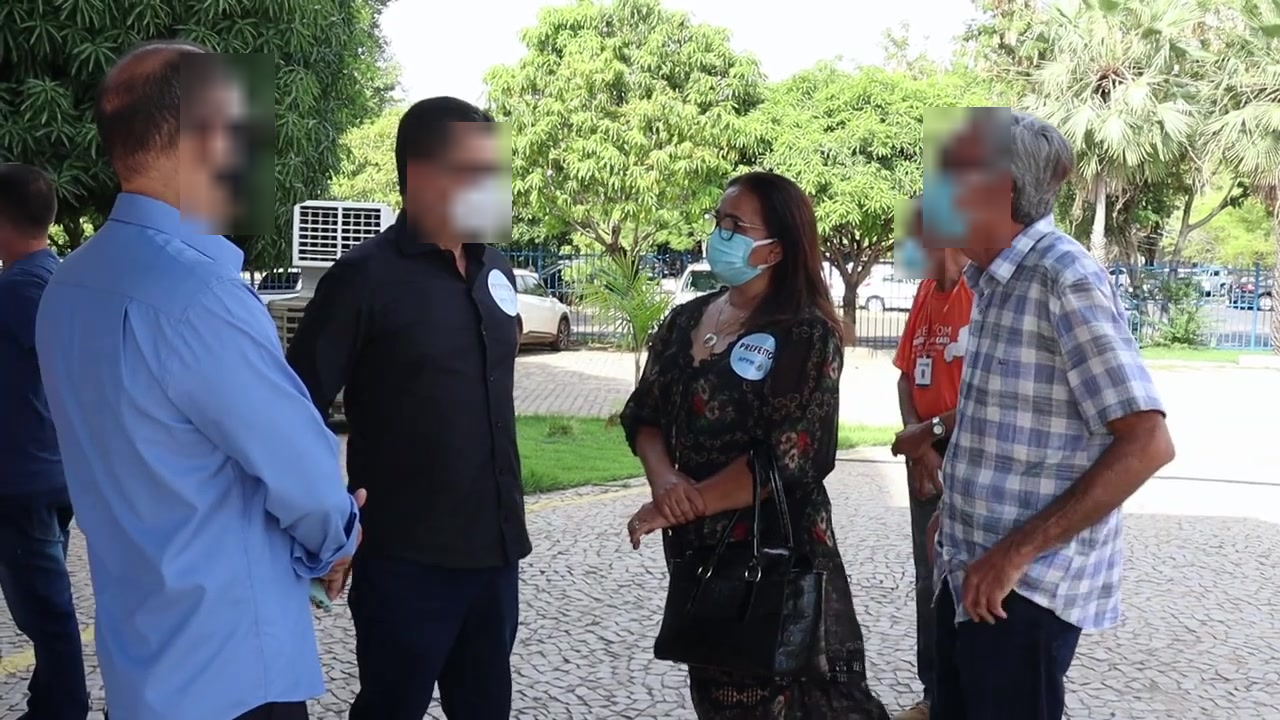}\hspace{0.5em}%
            \includegraphics[height=15.8ex]{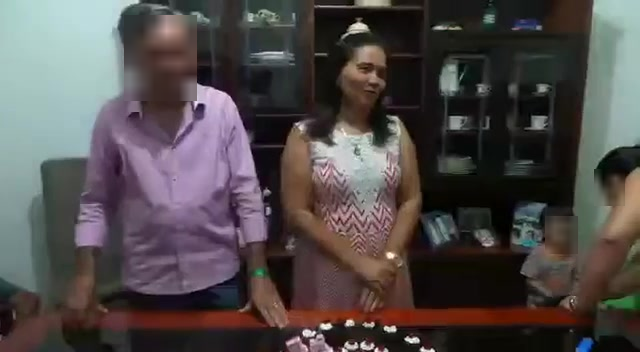}\hspace{0.5em}%
            \includegraphics[height=15.8ex]{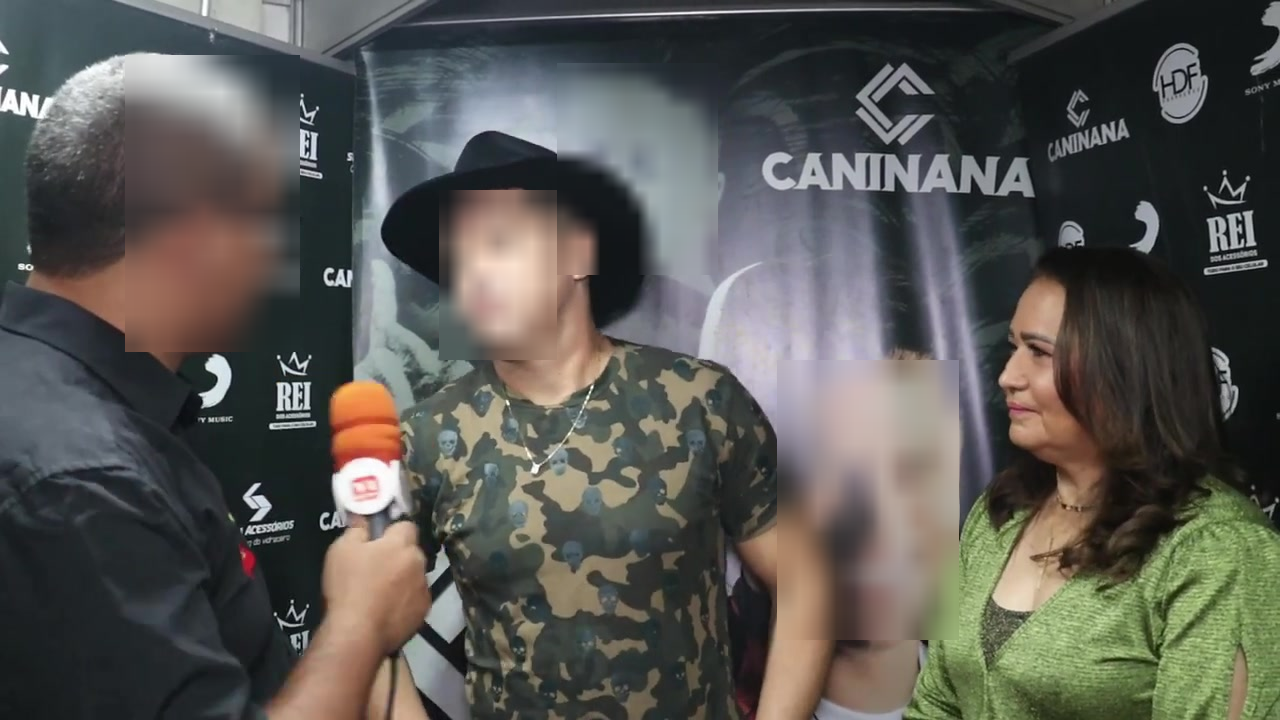}%
        }%
    }
    
    \resizebox{0.90\linewidth}{!}{%
        \subfloat[]{%
            \includegraphics[height=15.8ex]{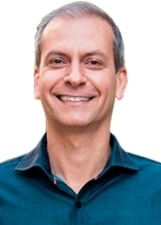}\hspace{0.5em}%
            \includegraphics[height=15.8ex]{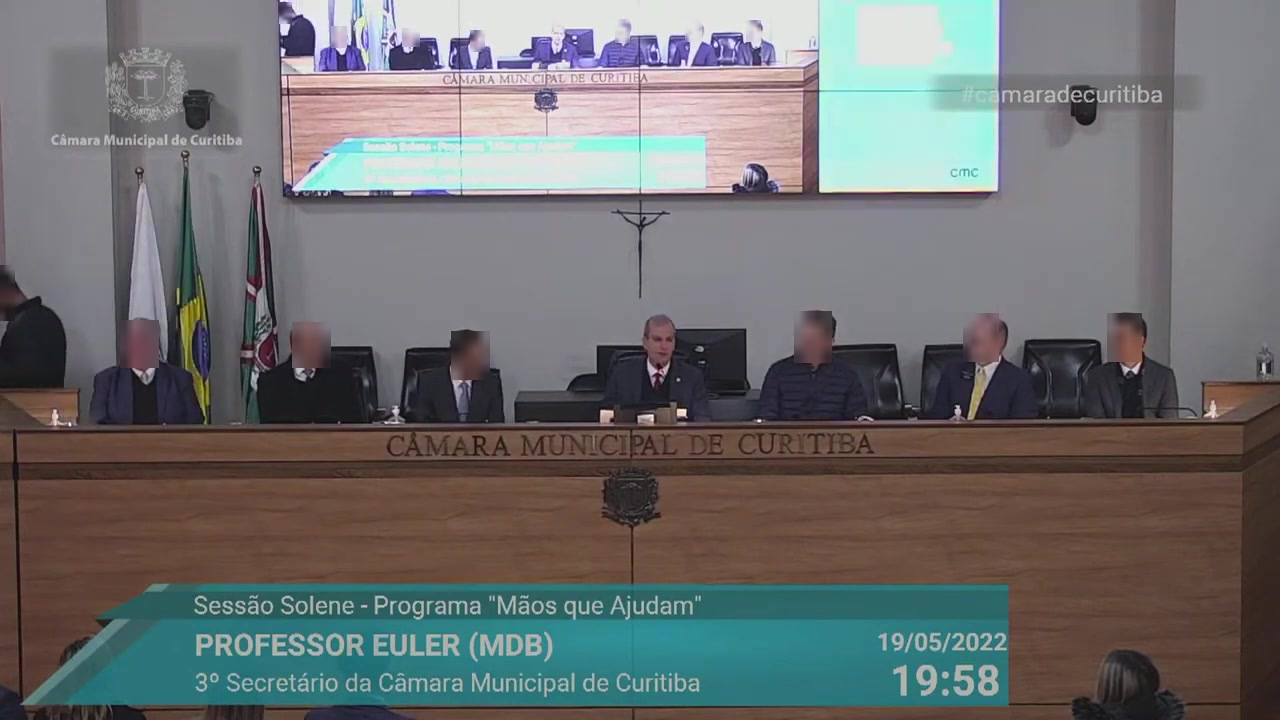}\hspace{0.5em}%
            \includegraphics[height=15.8ex]{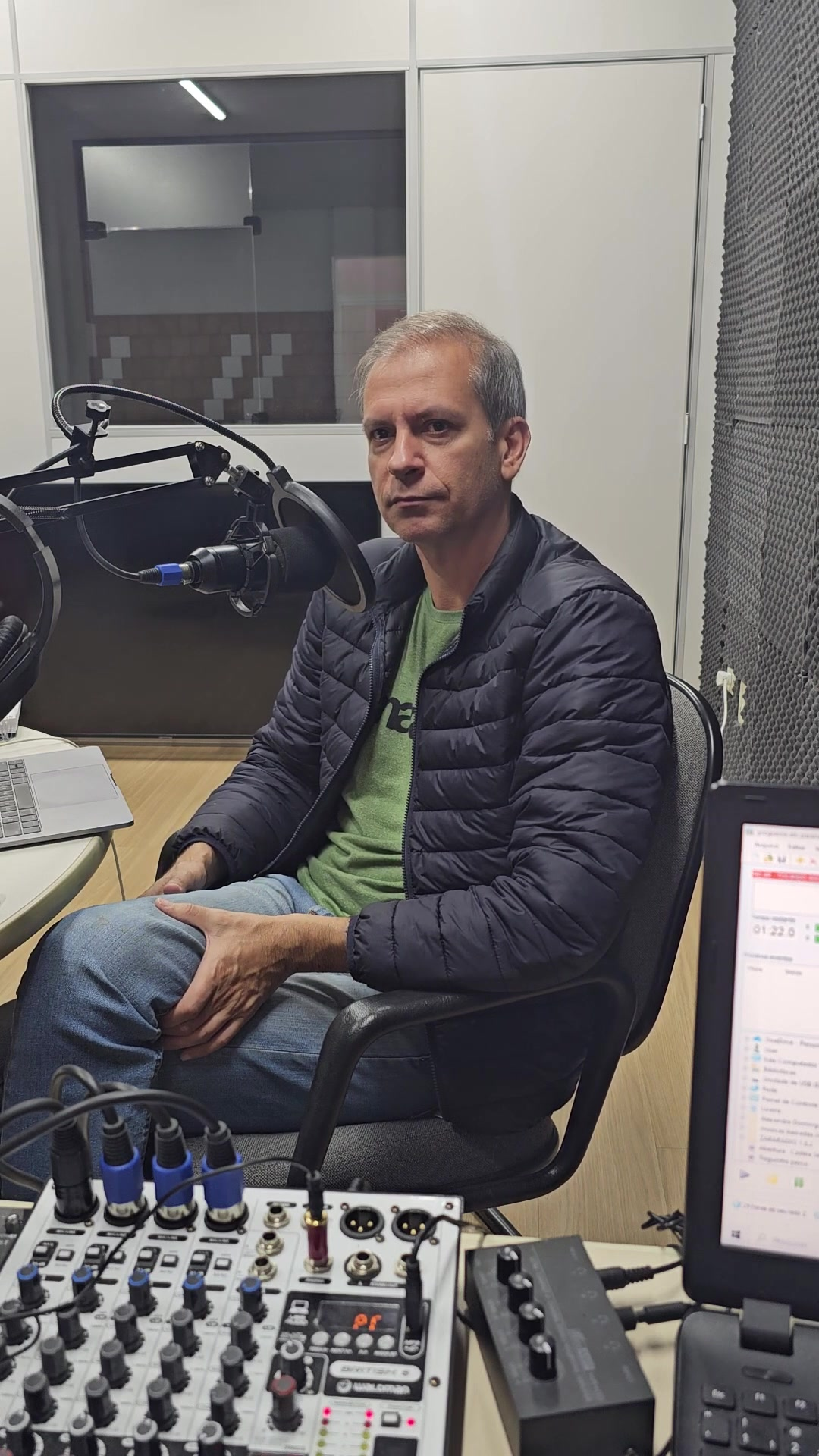}\hspace{0.5em}%
            \includegraphics[height=15.8ex]{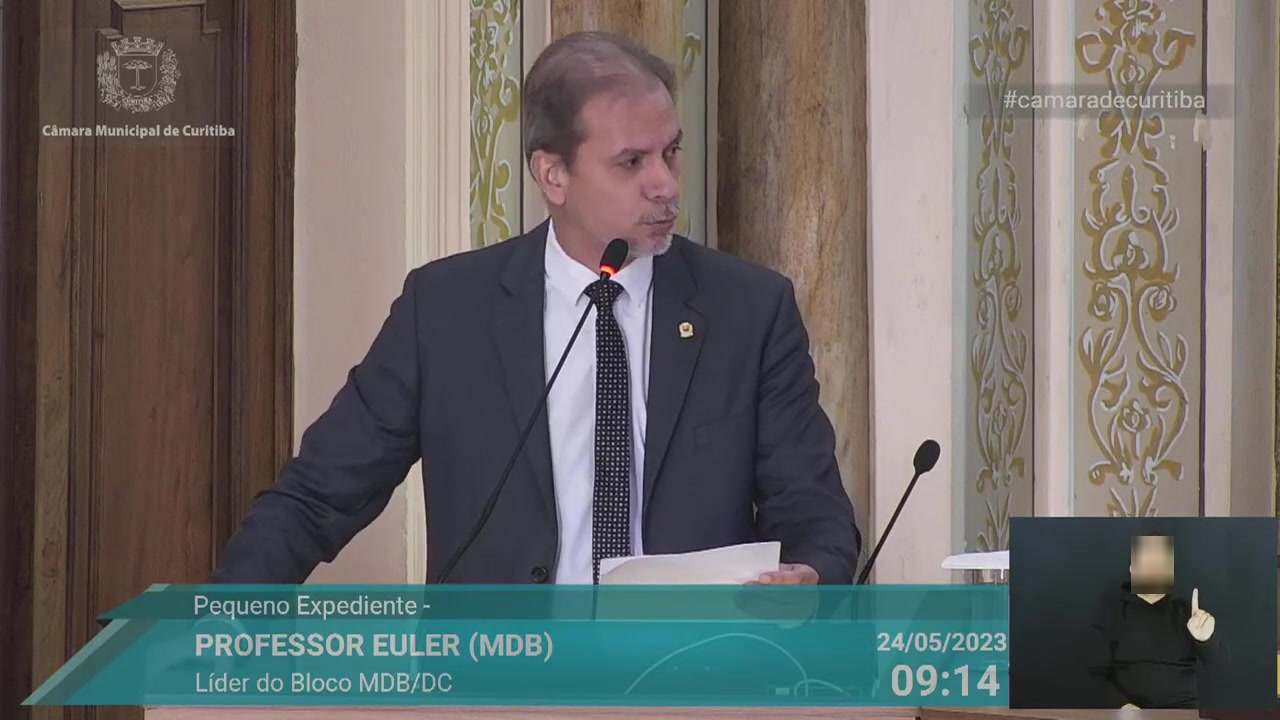}%
        }%
    }
    
    \caption{Representative samples from \dataset, illustrating the compressed and uncontrolled visual conditions present in the benchmark. Each row corresponds to a different politician; the first column shows the gallery image, while the remaining columns show frames extracted from public videos. Incidental faces were anonymized when necessary.}
    \label{fig:dataset_examples}
\end{figure}

Motivated by these limitations, this paper introduces \dataset\footnote{How to request the dataset and the image encoding, benchmark tests, and the code %
are available at: \href{https://github.com/UFPR-IPASP-PR/UFPR-PEs}{https://github.com/UFPR-IPASP-PR/UFPR-PEs}.}, a benchmark for face recognition bias evaluation under publicly available videos. The dataset is built from public videos of elected Brazilian politicians and annotated with self-declared race/color labels from official records. Although the proposed dataset is imbalanced across demographic dimensions, its composition reflects the actual availability of public records and source videos, making it a realistic benchmark for fairness analysis. Unlike many prior fairness benchmarks, our proposal uses official demographic declarations rather than third-party annotations or automatic labeling, thereby reducing label noise and aligning the benchmark with a legally grounded demographic source.

As illustrated in Fig.~\ref{fig:dataset_examples}, \dataset is designed to study whether demographic gaps become more pronounced under challenging visual conditions. This benchmark is motivated by the Brazilian demographic context; according to the 2022 Census~\cite{ibge2022censo}, the population is distributed as 45.3\% \textit{parda}, 43.5\% \textit{white}, 10.2\% \textit{black}, \ADdel{0.8\%}\AD{0.6\%} \textit{indigenous}, and 0.4\% \textit{yellow}. The \textit{parda} category is especially relevant because it has no direct equivalent in the U.S.- or Europe-centric taxonomies commonly used in prior face recognition benchmarks. \AD{We retain the Portuguese term rather than rendering it as \textit{mixed-race} or \textit{multiracial}, since those labels presuppose mixed ancestry, whereas the Brazilian category is self-declared and grounded in phenotype and social perception rather than in descent.\footnote{\AD{Unlike ancestry-based schemas, the Brazilian classification admits that individuals of the same descent may legitimately declare different categories, and it carries no genealogical claim~\cite{nogueira1985, telles2004}. No English label preserves this semantics, so we keep the original term throughout.}}}

To support this analysis, the benchmark preserves difficult samples rather than filtering them out, and it is paired with an evaluation protocol designed to capture both the data's difficulty and \ADdel{the persistence of demographic disparities across model families}\AD{how demographic disparities vary across difficulty strata}. The main contributions of this paper are: (1) \dataset is, to the best of our knowledge, the first face recognition bias benchmark to use self-declared race/color labels from official electoral records at scale; (2) it provides Brazilian and broader Latin American demographic coverage, including the \textit{parda} category absent from prior schemas; (3) it is built from compressed public video under uncontrolled conditions rather than curated still imagery; and (4) it supports stratified analysis by race/color and difficulty level, which is essential to studying how fairness gaps evolve across visual quality strata.

The remainder of the paper is organized as follows. Section~\ref{sec:related_work} reviews related work. Section~\ref{sec:benchmark} presents the benchmark and construction pipeline. Section~\ref{sec:results} reports the results of the fairness evaluation. Section~\ref{sec:limitations} discusses limitations, and Section~\ref{sec:conclusion} concludes the paper.

\begin{table*}[!t]
    \centering
    \caption{Comparison of representative face recognition bias benchmarks. Construction-evaluation overlap refers to whether the benchmark was built using the same or a closely related model family later used in evaluation.}
    \label{tab:related_work}
    \small
    \setlength{\tabcolsep}{3.5pt}
    \renewcommand{\arraystretch}{1.08}
    \begin{tabularx}{\textwidth}{l c c >{\centering\arraybackslash}X >{\centering\arraybackslash}X >{\centering\arraybackslash}X >{\centering\arraybackslash}X}
    \toprule
    Dataset & Identities & Images & \makecell{Demographic\\Axes} & \makecell{Annotation\\Source} & \makecell{Visual\\Conditions} & \makecell{Construction-Eval\\Overlap} \\
    \midrule
    RFW & ~$12{,}000$ & ~$\phantom{0,0}40{,}000$ & Race & Reorganized public data & Curated still images & Partial \\
    BFW & ~$\phantom{00,}800$ & ~$\phantom{0,0}20{,}000$ & Race, gender & Curated public data & Still images & Partial \\
    DemogPairs & ~$\phantom{00,}600$ & ~$\phantom{0,0}10{,}800$ & Race, gender & Dataset reorganization & Still images & Partial \\
    BUPT-GlobalFace & ~$38{,}000$ & ~$2{,}000{,}000$ & Race & Source metadata / API & Curated still images & Partial \\
    \dataset (ours) & ~$5{,}103$ & \ADdel{~$\phantom{0,}547{,}549$}\AD{~$\phantom{0,}547{,}519$} & Race/color, age, gender & Official election records & Compressed public video & \ADdel{None by design}\AD{Partial (reviewed)} \\
    \bottomrule
    \end{tabularx}
\end{table*}

\section{Related Work}
\label{sec:related_work}

Face recognition fairness benchmarks have evolved considerably in terms of demographic coverage, annotation methodology, and visual conditions. This section reviews the most representative datasets and identifies the limitations that \dataset is designed to address.

\subsection{Bias Benchmarks}

Several datasets have shaped fairness evaluation in face recognition~\cite{kotwal2025surveyX}. RFW~\cite{wang2019rfw} popularized race-aware benchmarking by reorganizing existing public data into racial subgroups for verification evaluation. BFW~\cite{robinson2020bfw} extended subgroup evaluation with a balanced verification design across race and gender. DemogPairs~\cite{hupont2019demogpairs} quantified demographic effects using pair-based analysis and showed that demographic imbalance in training data affects recognition accuracy across groups. BUPT-GlobalFace and BUPT-BalancedFace~\cite{buptdatasets} further explored how training composition affects subgroup behavior at scale. These datasets were important contributions, but they differ substantially from \dataset in annotation source, visual conditions, and construction methodology, as summarized in Table~\ref{tab:related_work}.

\subsection{Benchmark Limitations}

Prior benchmarks are limited in four ways. First, demographic labels are often produced by annotators or classifiers rather than drawn from self-declared records, introducing noise and subjective bias into the evaluation ground truth. Second, many datasets rely on curated still images that do not reflect the compression, blur, motion artifacts, and pose variation present in real-world video. Third, geographic coverage is narrow, with most benchmarks centered on U.S.- or Europe-centric demographic categories that do not transfer well to other contexts. Fourth, benchmark construction and evaluation can become circular when the same model family influences both stages, artificially inflating reported performance. \dataset is designed to avoid these problems by using self-declared labels from official records, compressed public video, Brazilian demographic categories, and zero-overlap evaluation.

\section{The \dataset Benchmark}
\label{sec:benchmark}

\dataset is a benchmark built from publicly available election-related videos designed to evaluate face recognition performance under challenging visual conditions, including variations in illumination, compression, blur, pose, expression, and occlusion. The current version is organized into two sets: \textit{gallery} and \textit{query}. 
The gallery comprises official portraits submitted during registration with the Brazilian Superior Electoral Court (TSE), featuring one distinct individual per image.
The query set consists of downloaded videos from a subset of these identities, which constitute the probe subjects, while the remaining gallery identities act as distractors during identification. 
Detailed dataset statistics are summarized in Table~\ref{tab:stats}.
All query images obtained from the videos were manually verified through semi-automated visual analysis by some of the authors of this work.

\begin{figure}[!tb]
    \centering
    \includegraphics[width=\columnwidth]{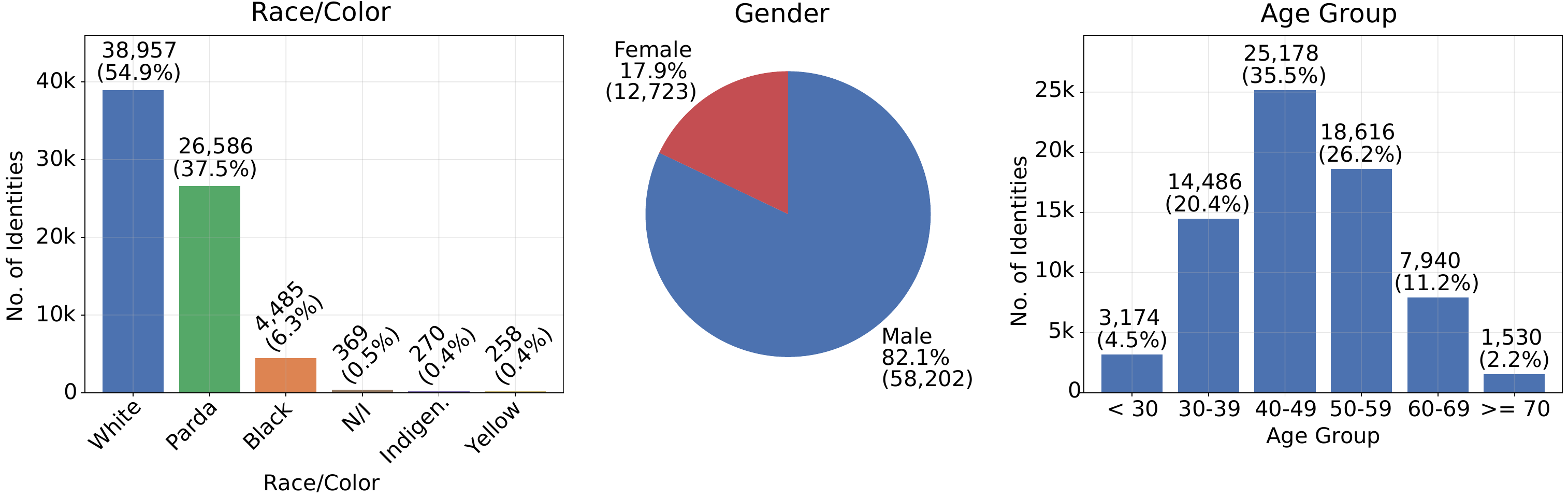}
    \caption{Distribution of \dataset by \ADdel{ethnic group}\AD{race/color group}, gender, and age.}
    \label{fig:distrib_ethnic_gender_age}
\end{figure}

\begin{table}[!b]
    \centering
    \caption{\dataset dataset statistics, detailing the composition of both gallery and query sets.}
    \label{tab:stats}
    \begin{tabular}{lc}
        \toprule
        Statistic & Value \\
        \midrule
        Gallery images & 70,861 \\
        Query identities & 5,103 \\
        Downloaded videos & 10,811 \\
        Video length per sample & 30 seconds \\
        Query images & \ADdel{547,549}\AD{547,519} \\
        Videos per identity & 2.12 \\
        \bottomrule
    \end{tabular}
\end{table}

To build the query set, each source video is trimmed to a 30-second segment to standardize temporal coverage and reduce the influence of very long clips. The input media is typically in the 360p--720p resolution range and encoded with H.264, introducing compression artifacts, motion blur, and pose variation that make the benchmark more challenging than curated face-image datasets. Rather than filtering difficult samples out, \dataset preserves them so that robustness and fairness can be evaluated under realistic conditions.

\subsection{Dataset Statistics}

All identities possess self-declared attribute labels for \ADdel{ethnic group}\AD{race/color group}, gender, and age. This characteristic significantly reduces annotation noise and grounds the benchmark in legal and administrative records rather than predictive models. Fig.~\ref{fig:distrib_ethnic_gender_age} summarizes the distribution across these attributes. 
\ADdel{Ethnic groups}\AD{Race/color groups} follow the official racial/color categories defined by the Brazilian Institute of Geography and Statistics (IBGE). Regarding age, the dataset distribution reflects the requirements of Brazilian electoral law: since only individuals aged 18+ are eligible to run for political office, our subjects are strictly restricted to this demographic, excluding younger age groups.

\subsection{Construction and Annotation}
\label{sec:benchmark:construction}

\begin{figure}[!b]
    \centering
    \includegraphics[width=0.78\columnwidth]{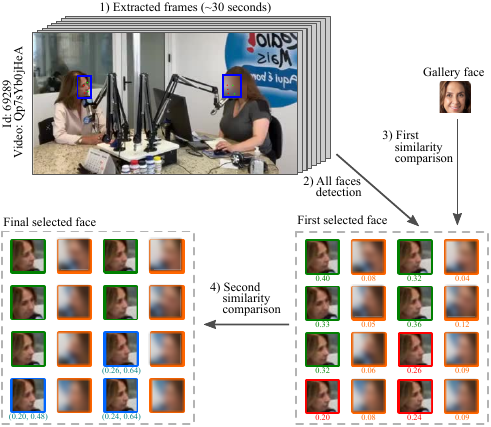}
    \caption{Illustration of our automatic gallery face selection process. In a first step, most of the target faces were correctly selected (green) and all other faces discarded (orange), based on their cosine similarity to the gallery. Observing that some challenging target faces were incorrectly discarded (red), we re-evaluated them against the gallery, at a lower threshold, and against previously selected faces (green), recovering most of the correct target faces (blue).}
    \label{fig:gallery_construction}
\end{figure}

Our benchmark is built through a reproducible video-to-identity pipeline. In each extracted video frame, we detect all faces using RetinaFace~\cite{Deng2020CVPR} and compare them to the gallery faces using cosine similarity, retaining only faces with similarity $\geq 0.3$ as the target politician. Because some challenging target faces may be missed in this first pass, a second selection step is applied with a lower gallery threshold of 0.2, recovering faces with an average similarity $\geq 0.25$ with respect to the previously selected frames. Fig.~\ref{fig:gallery_construction} presents an illustration of such a selection process, where the single real numbers \ADdel{bellow} \AD{below} correspond to the cosine similarity to the gallery, and number pairs correspond to the similarity to gallery and the average similarity to previously selected faces. The target correctly selected faces are surrounded in green, while the incorrectly discarded target faces are in red, the correctly discarded faces in orange, and the target recovered faces in blue. \ADdel{The automatically selected samples are then manually reviewed, and incorrect face selections are corrected when necessary.}\AD{Every extracted frame is then manually reviewed, regardless of the automatic decision: annotators inspect the gallery portrait alongside all frames of the video and mark both incorrect selections and target faces that were incorrectly discarded, so that false positives are removed and false negatives are recovered.} All non-target faces are blurred in the video frames after matching to preserve the identities of incidental~individuals.

To support stratified evaluation, each probe frame is assigned to one of three difficulty subsets based on the cosine similarity between its embedding and the corresponding gallery embedding: \textit{easy} (CS~$\geq 0.60$), characterized by high-quality frontal faces; \textit{medium} ($0.30 <$ CS~$< 0.60$), representing ambiguous cases with partial occlusion or moderate pose variation; and \textit{hard} (CS~$\leq 0.30$), containing low-quality, heavily occluded, or profile faces. These thresholds are consistent with values commonly adopted in the face recognition literature for ArcFace-based systems~\cite{deng2019arcface}.

\subsection{Access Policy and Ethics}

The \dataset dataset is restricted to academic purposes only and is available upon request\footnote{\iffinal\url{https://github.com/UFPR-IPASP-PR/UFPR-PEs}\else \url{github.com/anonymous}\fi}. Its construction was approved as a research project by the Ethics Committee Board from the \iffinal Human and Social Science Sector of the Federal University of Paraná, Brazil (Process CAAE 93419925.4.0000.0214)\else Anonymous Institution\fi, registered in the \href{https://plataformabrasil.saude.gov.br}{Plataforma Brasil} system.
While we obtained authorization to collect, handle, and analyze the publicly available data of the Brazilian politicians, we do not have the rights to publicly display the full dataset. Furthermore, the individuals whose images appear in this paper provided written informed consent to have their images used for publication purposes.

The benchmark is designed for research on face recognition fairness and robustness under public video conditions. Its release policy and final ethics approval status are documented separately, together with the institutional requirements for~access.

\section{Results}
\label{sec:results}

\begin{figure*}[!tb]
    \centering
    \includegraphics[width=0.9\textwidth]{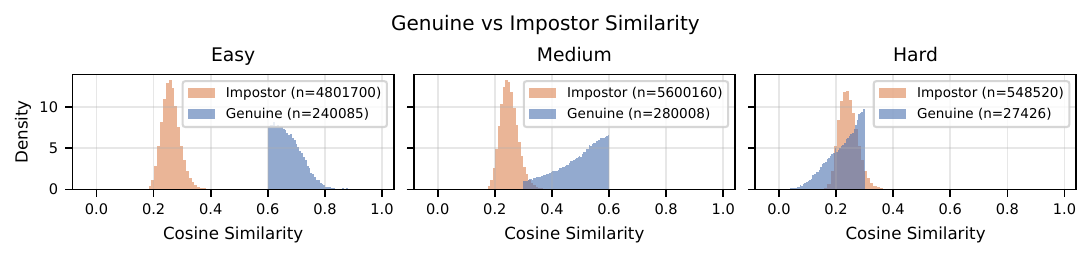}
    \vspace{-10pt}
    \caption{Cosine similarities between impostor and genuine pairs in the three subsets: easy, medium, and hard.}
    \label{fig:cosine_similarities_easy_medium_hard}
\end{figure*}

This section reports the main face recognition results obtained on \dataset. We evaluate verification (1:1) and identification (1:N) performance across the three difficulty subsets and then examine how recognition accuracy varies among \ADdel{ethnic groups}\AD{race/color} in both closed-set and open-set settings. 

To investigate these characteristics, we evaluated a ResNet50 (R50) model~\cite{he2016residual} trained with ArcFace loss~\cite{deng2019arcface} and pretrained on WebFace260M~\cite{Zhu2021WebFace260MAB}. Given an aligned and cropped input face image $i$ of size $112 \times 112$ pixels, normalized from $[0, 255]$ to $[-1, 1]$, R50 produces a face embedding $e_i \in \mathbb{R}^{512}$, which is compared to other embeddings using cosine similarity, i.e., 
\begin{equation}
    \mathrm{CS}(e_1, e_2) = \frac{e_1 \cdot e_2}{\left\| e_1 \right\| \left\| e_2 \right\|},
\label{eq:cosine_similarity_normalized}
\end{equation}
\noindent subject to $\left\| e_1 \right\| = \left\| e_2 \right\| = 1$.
Note that all experiments use the cosine similarity as the comparison metric.

The impact of visual degradation across the difficulty subsets is clearly reflected in the score distributions (Fig.~\ref{fig:cosine_similarities_easy_medium_hard}). 
While the high-quality frontal faces in the \textit{easy} subset yield near-perfect separability between genuine and impostor pairs, this margin decreases in the \textit{medium} subset. In contrast, the severe occlusions, low resolution, and profile poses present in the \textit{hard} subset cause a dramatic drop in genuine similarity scores, resulting in a large distribution overlap with impostor pairs.

\subsection{Overall Performances}

\begin{figure}[!tb]
    \centering
    \subfloat[Verification ROC by difficulty.]{%
        \includegraphics[width=0.7\columnwidth]{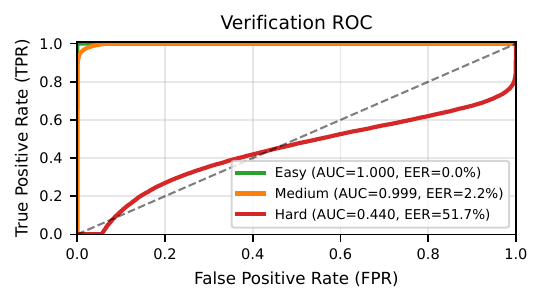}%
        \label{fig:roc_by_difficulty}}
    \\
    \subfloat[Closed-set CMC by difficulty.]{%
        \includegraphics[width=0.7\columnwidth]{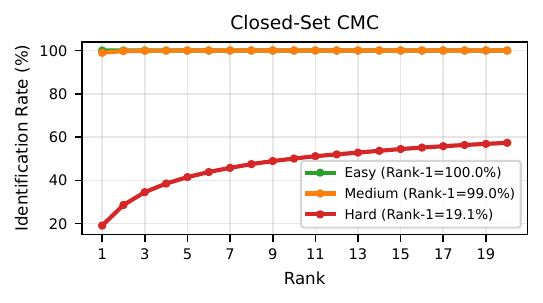}%
        \label{fig:cmc_by_difficulty}}
    \caption{Verification and identification performance of one R50+ArcFace model across the three difficulty subsets on \dataset. (a) ROC curves for the verification task (1:1); (b) CMC curves for closed-set identification (1:N).}
    \label{fig:roc_cmc_by_difficulty}
\end{figure}

In the verification (1:1) task, we generated 547,519 genuine pairs and 10,950,380 impostor pairs, for a total of \ADdel{11,487,899} \AD{11{,}497{,}899} pairs. The FR model must correctly predict whether two face images belong to the same subject. To evaluate verification performance, we compute the Area Under the Curve (AUC) of the Receiver-Operating Characteristic (ROC) curve, which describes the relationship between False Acceptance Rate (FAR) and True Acceptance Rate (TAR) as the similarity threshold decreases from 1 to 0. Fig.~\ref{fig:roc_by_difficulty} shows the ROC curves for R50 in the easy, medium, and hard subsets. Owing to their separability in feature space, the AUCs are 1.000 for easy, \ADdel{0.990}\AD{0.999} for medium, and \ADdel{0.540}\AD{0.440} for hard, \ADdel{indicating that the hardest condition remains substantially more challenging.} \AD{indicating a collapse of verification performance under severe degradation, with the hard subset falling below chance level.}

\begin{table}[!b]
    \centering
    \caption{Closed-set identification (1:N) performances on the subsets \textit{easy}, \textit{medium}, and \textit{hard}.}
    \label{tab:overall_performances}
    \begin{tabular}{lrrrr}
    \toprule
    Set & Total & Rank-1 (\%) & Rank-3 (\%) & Rank-5 (\%) \\
    \midrule
    Easy    & 240,085 & 99.97 & 100.00 & 100.00 \\
    Medium  & 280,008 & 98.97 & 99.91  & 99.98 \\
    Hard    &  27,426 & 19.10 & 34.55  & 41.49 \\
    Overall & 547,519 & 95.41 & 96.67  & 97.06 \\
    \bottomrule
    \end{tabular}
\end{table}

We also evaluated closed-set identification (1:N), where each query image belongs to a preexisting subject in the gallery and the model must assign the most similar identity. In this setting, R50 reached 99.97\%, 98.97\%, and 19.10\% rank-1 accuracy in the easy, medium, and hard subsets, respectively, as shown in Table~\ref{tab:overall_performances}. 
The corresponding rank-3 and rank-5 metrics confirm that performance drops sharply in the hard subset — even at rank-5, only 41.49\% of hard samples are correctly identified, demonstrating the severity of this operating condition.

\begin{figure}[!t]
    \centering
    \includegraphics[width=0.6\linewidth]{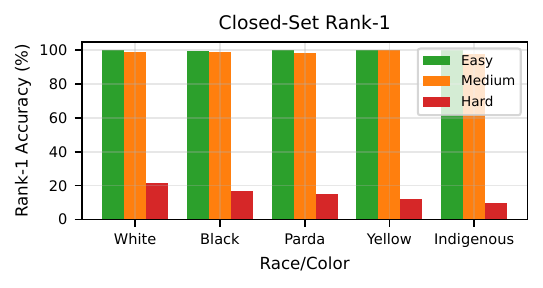}%
    \vspace{-10pt}
    \caption{Performance analysis of the model on the \dataset benchmark. Closed-set rank-1 accuracy of the model by race/color and difficulty subset.}
    \label{fig:rank1_acc_by_race_color}%
\end{figure}

\subsection{Bias by \ADdel{Ethnic Groups}\AD{Race/Color Groups}}

In addition to overall recognition performance, we analyzed closed-set and open-set identification accuracy across \ADdel{ethnic groups}\AD{race/color groups} and difficulty levels. As shown in Fig.~\ref{fig:rank1_acc_by_race_color}, the easy and medium subsets achieve very high rank-1 accuracy across all groups in the closed-set scenario, approaching perfect recognition. In the hard subset, however, performance drops substantially across all groups, revealing a much more challenging setting.

\begin{figure}[!tb]
    \centering
    \includegraphics[width=0.7\linewidth]{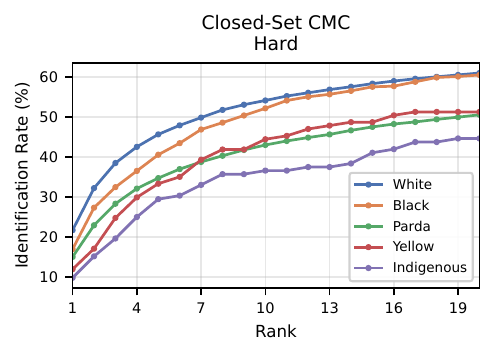}
    \vspace{-10pt}
    \caption{Closed-set identification (1:N) performance of a single R50+ArcFace model by \ADdel{ethnic group}\AD{race/color group}, shown as CMC curves on the \textit{hard} set.}      %
    \label{fig:rank1_cmc_race_by_race_hard}        
\end{figure}

Fig.~\ref{fig:rank1_cmc_race_by_race_hard} shows the CMC curves for closed-set identification on the hard subset, illustrating how identification rates improve as rank increases up to Rank-20.
Although expanding the candidate list size noticeably improves accuracy for all demographics, the curves reveal a persistent performance gap among \ADdel{ethnic groups}\AD{race/color groups} under these severe visual degradations. Specifically, the \textit{white} demographic maintains the highest identification rates across all ranks, followed by the \textit{black} and \textit{parda} groups, which exhibit very similar trajectories. Conversely, the model yields lower cumulative accuracy for the \textit{yellow} and \textit{indigenous} categories, highlighting demographic disparities that are heavily pronounced in challenging operational conditions.

\begin{figure*}[!t]
    \centering
    \includegraphics[width=0.98\textwidth]{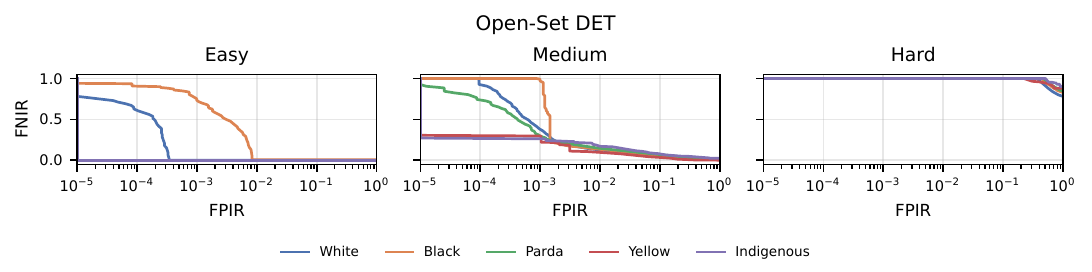}
    \vspace{-10pt}
    \caption{Open-set identification analysis of the model R50+ArcFace by race/color and difficulty subsets on our \dataset benchmark.}
    \label{fig:openset_det}%
\end{figure*}

We also evaluated open-set identification, where the correct identity of a query face may or may not exist in the gallery, and the model must distinguish between known and unknown identities. Performances were analyzed by comparing the False Negative Identification Rate (FNIR) and False Positive Identification Rate (FPIR). In the open-set setting, the groups show moderate performance differences, as shown in Fig.~\ref{fig:openset_det}. 
The \textit{indigenous} group attains the highest rank-1 accuracy, whereas the \textit{yellow} group shows the lowest result, and the remaining groups cluster around the overall average. This pattern is consistent across difficulty levels, with the performance gap between groups widening in the hard subset. 

\section{Limitations}
\label{sec:limitations}

\dataset has several limitations that should be considered when interpreting its statistics and experimental results.

First, the benchmark is constructed from self-declared demographic records and therefore reflects a social and administrative classification rather than a biological one. Although this choice reduces annotation noise and grounds the labels in an official source, it also inherits any inconsistencies present in the records themselves. In particular, self-declared racial categories may be affected by the institutional and political context in which politicians register, including policies related to electoral quotas and public funding. As a result, the labels should be interpreted as administrative self-declarations rather than fixed biological categories.

Second, the benchmark is built from public political videos, so its demographic and regional composition is shaped by the availability of online material and by the election context. Because the benchmark is built from publicly available political videos, the amount and quality of material can vary across regions and groups. This may affect both the size of the probe sets and the distribution of difficulty levels across identities. As a result, some groups may remain underrepresented even after stratified sampling.

\AD{Third, the query set is initially assembled by an automated gallery search, so the model used during construction produces the first \textit{selected} / \textit{discarded} decision for each extracted frame. This decision is not final: as described in Section~III-B, annotators exhaustively review every extracted frame and correct both classes, which removes the false positives and recovers the false negatives introduced by the search. A residual dependence remains at the detection stage, since faces the detector fails to localize never reach either the search or the review. The direction of this residual effect is conservative with respect to our conclusions: undetected faces are concentrated in the most degraded observations, so the reported gaps are expected to underestimate rather than to overstate the disparity between race/color groups.}

\ADdel{Third}\AD{Fourth}, the current evaluation covers a single model family — R50 with ArcFace loss pretrained on WebFace260M. While this model represents a strong and widely used baseline, the observed demographic gaps may not generalize to other architectures, loss functions, or training datasets. Future work should extend the evaluation to additional models to assess whether the reported disparities are model-specific or inherent to the benchmark conditions.

\section{Conclusion}
\label{sec:conclusion}

We introduced \dataset, a benchmark for face recognition bias evaluation built from compressed public video of elected Brazilian politicians, annotated with official self-declared race/color labels from TSE records. The dataset combines demographically grounded annotations, uncontrolled visual conditions, and explicit difficulty stratification to study whether demographic gaps persist and vary across recognition operating conditions.

Our experimental results show that subgroup performance must be interpreted jointly with image quality, since the gaps are not constant across difficulty levels. In the easy and medium subsets, recognition performance is high across all \ADdel{ethnic groups}\AD{race/color groups}. In the hard subset, performance drops substantially for every group, and the relative differences between groups widen. This reinforces the need for benchmarks that preserve difficult samples instead of filtering them out, and suggests that model comparisons can obscure subgroup-specific behavior when quality variation is ignored.

Overall, \dataset provides a reproducible and demographically grounded setting for studying face recognition bias under realistic public video conditions, with coverage of the Brazilian \textit{parda} category that is absent from prior benchmarks. Future work will expand the evaluation to additional model families, refine the analysis under alternative operating points, and investigate how the observed disparities evolve across different training data compositions.

\section*{\uppercase{Acknowledgments}}

\iffinal
    This study was financed in part by the \textit{Coordenação de Aperfeiçoamento de Pessoal de Nível Superior - Brasil~(CAPES)}, through the \textit{Programa de Excelência Acadêmica~(PROEX)} - Finance Code 001, and in part by the \textit{Conselho Nacional de Desenvolvimento Científico e Tecnológico~(CNPq)} and \textit{Fundação Araucária} under grant \#078/2026. The authors also thank the Federal Institute of Mato Grosso (IFMT), Pontes e Lacerda Campus, for supporting Bernardo Biesseck.
\else
    The acknowledgments are hidden for review.
\fi

\iffinal
 \balance  %
\else
\fi

\bibliographystyle{IEEEtran}
\bibliography{bibtex}

@inproceedings{buolamwini2018gender,
  author    = {Buolamwini, Joy and Gebru, Timnit},
  title     = {Gender Shades: Intersectional Accuracy Disparities in Commercial Gender Classification},
  booktitle = {1st Conference on Fairness, Accountability and Transparency},
  year      = {2018},
  pages     = {77--91}
}

@inproceedings{raji2019actionable,
  author    = {Raji, Inioluwa Deborah and Buolamwini, Joy},
  title     = {Actionable Auditing: Investigating the Impact of Publicly Naming Biased Performance Results of Commercial {AI} Products},
  booktitle = {AAAI/ACM Conference on AI, Ethics, and Society},
  year      = {2019},
  pages     = {429--435}
}

@inproceedings{deng2019arcface,
  author    = {Deng, Jiankang and Guo, Jia and Xue, Niannan and Zafeiriou, Stefanos},
  title     = {ArcFace: Additive Angular Margin Loss for Deep Face Recognition},
  booktitle = {IEEE/CVF Conference on Computer Vision and Pattern Recognition (CVPR)},
  year      = {2019},
  pages     = {4690--4699}
}

@inproceedings{wang2019rfw,
  author    = {Wang, Mei and Deng, Weihong and Hu, Jiani and Tao, Xunqiang and Huang, Yaohai},
  title     = {Racial Faces in the Wild: Reducing Racial Bias by Information Maximization Adaptation Network},
  booktitle = {IEEE/CVF International Conference on Computer Vision (CVPR)},
  year      = {2019},
  pages     = {692--702}
}

@inproceedings{robinson2020bfw,
  author    = {Robinson, Joseph P. and Qin, Can and Henon, Yann and Timoner, Samson and Fu, Yun},
  title     = {Face Recognition: Too Bias, or Not Too Bias?},
  booktitle = {IEEE/CVF Conference on Computer Vision and Pattern Recognition Workshops (CVPRW)},
  year      = {2020}
}

@inproceedings{hupont2019demogpairs,
  author    = {Hupont, Isabel and Fern{\'a}ndez, Carles},
  title     = {DemogPairs: Quantifying the Impact of Demographic Imbalance in Deep Face Recognition},
  booktitle = {IEEE International Conference on Automatic Face and Gesture Recognition (FG)},
  year      = {2019}
}

@article{kotwal2025surveyX,
  author    = {Kotwal, Karan and Marcel, S{\'e}bastien},
  title     = {Fairness in Face Recognition: A Survey},
  journal   = {arXiv preprint arXiv:2502.02309},
  year      = {2025}
}

@misc{ibge2022censo,
  author       = {{Brazilian Institute of Geography and Statistics}},
  title        = {2022 Census},
  howpublished = {\url{https://www.ibge.gov.br/en/statistics/social/population/22836-2022-census-3.html}},
  year         = {2025},
  note         = {Accessed 2026-05-01}
}

@article{buptdatasets,
  title={Meta Balanced Network for Fair Face Recognition},
  author={Wang, Mei and Zhang, Yaobin and Deng, Weihong},
  journal={IEEE Transactions on Pattern Analysis and Machine Intelligence},
  year={2021},
  publisher={IEEE},
  note = {dataset available at \url{http://www.whdeng.cn/RFW/Trainingdataste.html}},
}

@inproceedings{Deng2020CVPR,
title = {RetinaFace: Single-Shot Multi-Level Face Localisation in the Wild},
author = {Deng, Jiankang and Guo, Jia and Ververas, Evangelos and Kotsia, Irene and Zafeiriou, Stefanos},
booktitle = {IEEE/CVF Conference on Computer Vision and Pattern Recognition (CVPR)},
year = {2020}
}

@inproceedings{he2016residual,
  author = {He, Kaiming and Zhang, Xiangyu and Ren, Shaoqing and Sun, Jian},
  title = {Deep Residual Learning for Image Recognition},
  booktitle = {IEEE/CVF Conference on Computer Vision and Pattern Recognition (CVPR)},
  year = 2016
}

@article{Zhu2021WebFace260MAB,
  title={WebFace260M: A Benchmark Unveiling the Power of Million-Scale Deep Face Recognition},
  author={Zheng Zhu and Guan Huang and Jiankang Deng and Yun Ye and Junjie Huang and Xinze Chen and Jiagang Zhu and Tian Yang and Jiwen Lu and Dalong Du and Jie Zhou},
  journal={2021 IEEE/CVF Conference on Computer Vision and Pattern Recognition (CVPR)},
  year={2021},
  pages={10487-10497},
  url={https://api.semanticscholar.org/CorpusID:232148002}
}

@book{nogueira1985,
  author    = {O. Nogueira},
  title     = {Tanto Preto Quanto Branco: Estudos de Relações Raciais},
  publisher = {T. A. Queiroz},
  address   = {São Paulo},
  year      = {1985}
}

@book{telles2004,
  author    = {E. E. Telles},
  title     = {Race in Another America: The Significance of Skin Color in Brazil},
  publisher = {Princeton University Press},
  address   = {Princeton, NJ},
  year      = {2004}
}

\end{document}